\pdfoutput=1

\documentclass{article}

\usepackage{times}
\usepackage{graphicx} 
\usepackage{subfig} 

\usepackage{natbib}

\usepackage{algorithm}
\usepackage{algorithmic}

\usepackage{graphicx} 
\usepackage{booktabs}       
\usepackage{amsfonts}       
\usepackage{nicefrac}       
\usepackage{microtype}      
\usepackage{caption}
\usepackage{amsmath}
\usepackage{amssymb}

\usepackage{bm}

\usepackage{hyperref}

\usepackage[accepted]{icml2017}

\icmltitlerunning{Uncertainty Decomposition in Bayesian Neural Networks}

\begin{document} 

\twocolumn[
\icmltitle{Uncertainty Decomposition in Bayesian Neural\\ Networks with Latent Variables}



\icmlsetsymbol{equal}{*}

\begin{icmlauthorlist}
\icmlauthor{Stefan Depeweg}{si,tum}
\icmlauthor{Jos\'e Miguel Hern\'andez-Lobato}{cam}
\icmlauthor{Finale Doshi-Velez}{ha}
\icmlauthor{Steffen Udluft}{si}
\end{icmlauthorlist}

\icmlaffiliation{si}{Siemens AG}
\icmlaffiliation{tum}{Technical University of Munich}
\icmlaffiliation{cam}{University of Cambridge}
\icmlaffiliation{ha}{Harvard University}

\icmlcorrespondingauthor{Stefan Depeweg}{stefan.depeweg@siemens.com}


\vskip 0.3in
]



\printAffiliationsAndNotice{}  

\begin{abstract} 
Bayesian neural networks (BNNs) with latent variables are probabilistic models
which can automatically identify complex stochastic patterns in the data. We
describe and study in these models a decomposition of predictive uncertainty into its
epistemic and aleatoric components. First, we show how such a decomposition
arises naturally in a Bayesian active learning scenario by following an
information theoretic approach. Second, we use a similar decomposition to
develop a novel risk sensitive objective for safe reinforcement learning (RL).
This objective minimizes the effect of model bias in environments whose
stochastic dynamics are described by BNNs with latent variables. Our
experiments illustrate the usefulness of the resulting decomposition in
active learning and safe RL settings.
\end{abstract}

\section{Introduction}

Recently, there has been an increased interest in Bayesian neural networks
(BNNs) and their possible use in reinforcement learning (RL) problems
\citep{gal2016improving,blundell2015weight,houthooft2016vime}. In particular,
recent work has extended BNNs with a latent variable model to describe complex stochastic
functions \citep{depeweg2016learning,moerland2017learning}. The proposed
approach enables the automatic identification of arbitrary stochastic patterns
such as multimodality and heteroskedasticity, without having to manually
incorporate these into the model. 

In model-based RL, the aforementioned BNNs with latent
variables can be used to describe complex stochastic dynamics. The BNNs encode
a probability distribution over stochastic functions, with each function
serving as an estimate of the ground truth continuous Markov Decision Process
(MDP). Such probability distribution can then be used for policy search, by finding the
optimal policy with respect to state trajectories simulated from the model. The BNNs with
latent variables produce improved probabilistic predictions and these result in better
performing policies \citep{depeweg2016learning,moerland2017learning}.

We can identify two distinct forms of uncertainties in the class of models
given by BNNs with latent variables. As described by \citet{kendall2017uncertainties},
"Aleatoric uncertainty captures noise inherent in the observations. On the
other hand, epistemic uncertainty accounts for uncertainty in the model." In
particular, epistemic uncertainty arises from our lack of knowledge of the values of
the synaptic weights in the network, whereas aleatoric uncertainty originates from
our lack of knowledge of the value of the latent variables.

In the domain of model-based RL the epistemic uncertainty
is the source of model bias (or representational bias, see e.g.
\citet{joseph2013reinforcement}). When there is high discrepancy between model
and real-world dynamics, policy behavior may deteriorate. In analogy to the
principle that "a chain is only as strong as its weakest link" a drastic error
in estimating the ground truth MDP at a single transition step can render the
complete policy useless (see e.g. \citet{schneegass2008uncertainty}).

In this work we address the decomposition of the uncertainty present in the
predictions of BNNs with latent variables into its epistemic and aleatoric
components. We show the usefulness of such decomposition in two different
domains: active learning and risk-sensitive RL.
 
First we consider an active learning scenario with stochastic functions. We
derive an information-theoretic objective that decomposes the entropy of the
predictive distribution of BNNs with latent variables into its epistemic and
aleatoric components. By building on that decomposition, we then investigate
safe RL using a risk-sensitive criterion \citep{garcia2015comprehensive} which
focuses only on risk related to model bias, that is, the risk of the policy
performing at test time significantly different from at training time. The
proposed criterion quantifies the amount of epistemic uncertainty (model bias risk)
in the model's predictive distribution and ignores any risk stemming from the
aleatoric uncertainty. Our experiments show that, by using this risk-sensitive
criterion, we are able to find policies that, when evaluated on the ground
truth MDP, are safe in the sense that on average they do not deviate
significantly from the performance predicted by the model at training time.

We focus on the off-policy batch RL scenario, in which we are
given an initial batch of data from an already-running system and are asked to
find a better policy. Such scenarios are common in real-world industry settings
such as turbine control, where exploration is restricted to avoid possible
damage to the system.

\def\H{\mathrm{H}}
\newcommand{\x}{\mathbf{x}}
\newcommand{\EI}{\textrm{EI}}
\newcommand{\C}{\mathcal{C}}
\newcommand{\given}{\,|\,}
\newcommand{\DistGam}{\text{Gam}}

\section{Bayesian Neural Networks with Latent Variables}\label{sec:bnnswlv}

Given data~${\mathcal{D} = \{ \mathbf{x}_n, \mathbf{y}_n \}_{n=1}^N}$, formed
by feature vectors~${\mathbf{x}_n \in \mathbb{R}^D}$ and targets~${\mathbf{y}_n
\in \mathbb{R}}^K$, we assume that~${\mathbf{y}_n =
f(\mathbf{x}_n,z_n;\mathcal{W}) + \bm \epsilon_n}$, where~$f(\cdot ,
\cdot;\mathcal{W})$ is the output of a neural network with weights
$\mathcal{W}$ and $K$ output units. The network receives as input the feature vector $\mathbf{x}_n$ and the
latent variable $z_n \sim \mathcal{N}(0,\gamma)$.
The activation functions for the hidden layers are rectifiers:~${\varphi(x) = \max(x,0)}$.
The activation functions for the output layers are the identity function:~${\varphi(x) = x}$.
The network output is
corrupted by the additive noise variable~$\bm \epsilon_n \sim \mathcal{N}(\bm 0,\bm
\Sigma)$ with diagonal covariance matrix $\bm \Sigma$. The role of the latent variable $z_n$ is to
capture unobserved stochastic features that can affect the network's
output in complex ways. Without $z_n$, randomness is only given by the additive
Gaussian observation noise $\bm \epsilon_n$, which can only describe limited stochastic
patterns. 
The network has~$L$ layers, with~$V_l$ hidden units in layer~$l$,
and~${\mathcal{W} = \{ \mathbf{W}_l \}_{l=1}^L}$ is the collection of~${V_l
\times (V_{l-1}+1)}$ weight matrices.  
The $+1$ is introduced here to account
for the additional per-layer biases. 
We approximate the exact posterior distribution $p(\mathcal{W},\mathbf{z}\given\mathcal{D})$
with the factorized Gaussian distribution
\begin{align}
q(\mathcal{W},\mathbf{z}) =  &\left[ \prod_{l=1}^L\! \prod_{i=1}^{V_l}\!  \prod_{j=1}^{V_{l\!-\!1}\!+\!1} \mathcal{N}(w_{ij,l}| m^w_{ij,l},v^w_{ij,l})\right] \\
&\left[\prod_{n=1}^N \mathcal{N}(z_n \given m_n^z, v_n^z) \right]\,.\label{eq:posterior_approximation}
\end{align}
The parameters~$m^w_{ij,l}$,~$v^w_{ij,l}$ and
~$m^z_n$,~$v^z_n$ are determined by
minimizing a divergence between 
$p(\mathcal{W},\mathbf{z}\given\mathcal{D})$
and the approximation $q$. For more detail the reader is referred to the work of 
\citet{hernandez2016black,depeweg2016learning}. In all our experiments we use
black-box $\alpha$-divergence minimization with $\alpha=1.0$, as it seems to
produce a better decomposition of uncertainty into its
empistemic and aleatoric components, although further studies
are needed to strengthen this claim. 

The described BNNs with latent variables can describe complex stochastic
patterns while at the same time account for model uncertainty. They achieve
this by jointly learning $q(\mathbf{z})$, which captures the specific values of
the latent variables in the training data, and $q(\mathcal{W})$, which
represents any uncertainty about the model parameters.  The result is a principled
Bayesian approach for inference of stochastic functions.

\section{Active Learning of Stochastic Functions}\label{sec:al}

Active learning is the problem of choosing which data points to incorporate
next into the training data so that the resulting gains in predictive
performance are as high as possible. In this section, we derive a
Bayesian active learning procedure for stochastic functions. This procedure
illustrates how to separate two sources of uncertainty, that is, aleatoric
and epistemic, in the predictive distribution of BNNs with latent variables.

Within a Bayesian setting,
active learning can be formulated as choosing data
based on the expected reduction in entropy of the posterior distribution
\cite{mackay1992information}. \citet{hernandez2015probabilistic} apply this
entropy-based approach to scalable BNNs. In \cite{houthooft2016vime}, the authors use
a similar approach as an exploration scheme in RL problems in which a BNN
is used to represent current knowledge about the transition dynamics. These
previous works only assume additive Gaussian noise and, unlike
the BNNs with latent variables from Section \ref{sec:bnnswlv},
they and cannot capture complex stochastic patterns.


We start by deriving the expected reduction
in entropy in BNNs with latent variables. We assume a scenario
in which a BNN with latent variables has been fitted to a batch of data
$\mathcal{D}=\{(\mathbf{x}_1,\mathbf{y}_i),\cdots,(\mathbf{x}_N,\mathbf{y}_N)\}$ to produce
a posterior approximation $q(\mathcal{W},\mathbf{z})$. We now want to estimate the expected reduction in
posterior entropy for $\mathcal{W}$ when a particular data point $\mathbf{x}$ is incorporated in the training
set. The expected reduction in entropy is
\begin{align}
  \text{H}(\mathcal{W}&|\mathcal{D}) - \mathbf{E}_{y|\mathbf{x},\mathcal{D}}\biggl[\text{H}(\mathcal{W}|\mathcal{D}\cup\{\mathbf{x},y\})\biggr] \\
  &= \text{H}(\mathcal{W})-\text{H}(\mathcal{W}|y)\\
  &= \text{I}(\mathcal{W};y) \\
  &=\text{H}(y)-\text{H}(y|\mathcal{W})\label{eq:intermediate}\\
  \begin{split}
   &= \text{H}\left[\int_{\mathcal{W},z}p(y|\mathcal{W},\mathbf{x},z)p(z)p(\mathcal{W}|\mathcal{D})dzd\mathcal{W}\right]  \\
   & - \mathbf{E}_{\mathcal{W}|\mathcal{D}} \biggl[  \text{H}(\int_z p(y|\mathcal{W}=\mathcal{W}_i,\mathbf{x},z) p(z) dz )\biggr] 
   \end{split} \label{eq:al-objective} 
\end{align}
where $\text{H}(\cdot)$ denotes the entropy of a random variable and
$\text{H}(\cdot|\cdot)$ and $\text{I}(\cdot;\cdot)$ denote the conditional
entropy and the mutual information between two random variables. In
(\ref{eq:intermediate}) and (\ref{eq:al-objective}) we see that the active
learning objective is given by the difference of two terms. The first term is
the entropy of the predictive distribution, that is, $\text{H}(y)$. The second
term, that is, $\text{H}(y|\mathcal{W})$, is a conditional entropy.
To compute this term, 
we average across $\mathcal{W}_i \sim q(\mathcal{W})$ 
the entropy
$\text{H}\left[\int_{z}p(y|\mathcal{W}=\mathcal{W}_i,\mathbf{x},z)p(z)dz\right]$.
As shown in this expression, the randomness in $p(y|\mathcal{W}=\mathcal{W}_i,\mathbf{x})$
has its origin in the latent variable $z$ (and the constant output noise
$\epsilon \sim \mathcal{N}(0,\bm \Sigma)$ which is not shown here). 
Therefore, this second term can be interpreted as the 'aleatoric uncertainty'
present in the predictive distribution, that is, the average entropy of $y$
that originates from the latent variable $z$ and not from the uncertainty about
$\mathcal{W}$. We can refer to the whole objective function in (\ref{eq:al-objective}) as an estimate of
the epistemic uncertainty: the full predictive uncertainty about $y$ given $\mathbf{x}$ minus the corresponding aleatoric
uncertainty.

\begin{figure}
\centering
\includegraphics[width=0.5\linewidth]{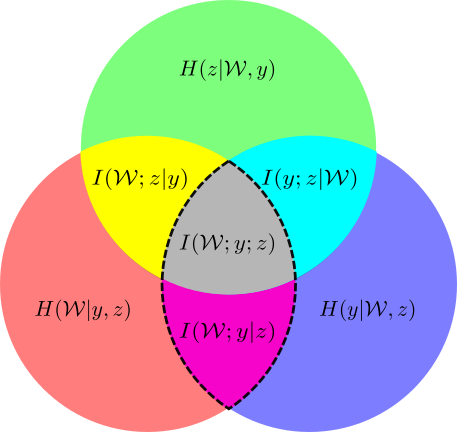}
\caption{Information Diagram illustrating quantities of entropy with three variables.
The area surrounded by a dashed line indicates reduction in entropy given by
equation (\protect{\ref{eq:al-objective}}). The conditioning to $\mathbf{x}$ is omitted for
readability.} \label{venn3d}
\end{figure}

The previous decomposition is also illustrated by the information diagram from Figure \ref{venn3d}. The
entropy of the predictive distribution is composed of the blue, cyan, grey and
pink areas. The blue area is constant: when both $\mathcal{W}$ and $z$ are
determined, the entropy of $y$ is constant and given by the entropy of the additive Gaussian noise $\epsilon \sim \mathcal{N}(0,\bm \Gamma)$.
$\text{H}(y|\mathcal{W})$ is given by the light and dark blue areas. The reduction in
entropy is therefore obtained by the grey and pink areas.

The quantity in equation (\ref{eq:al-objective}) can be approximating using standard entropy estimators, e.g. nearest-neighbor methods \citep{kozachenko1987sample,kraskov2004estimating,gao2016breaking}. For that, we 
repeatedly sample $\mathcal{W}$ and $z$ and do forward passes through the neural network to sample $y$.
The resulting samples of $y$ can then be used to
approximate the respective entropies for each $\mathbf{x}$ using the nearest-neighbor approach:
\begin{align}
&\text{H}(y|\mathbf{x},\mathcal{D})-\mathbf{E}_{W|\mathcal{D}}\left[\text{H}(\int_{z}p(y|\mathcal{W},\mathbf{x},z)p(z)dz)\right]\nonumber \\
&\approx \hat{\text{H}}(y_1,\ldots,y_L) -\frac{1}{M} \sum_{i=1}^M \left[\hat{\text{H}}(y_1^{\mathcal{W}_i},\ldots,y_L^{\mathcal{W}_i})\right]\,.  
\label{eq:entropy_after}
\end{align}
where $\hat{\text{H}}(\cdot)$ computes the nearest-neighbor estimate of the entropy
given an empirical sample of points, $y_1,\ldots,y_L\sim p(y|\mathbf{x},\mathcal{D})$,
$\mathcal{W}_1,\ldots,\mathcal{W}_M\sim q(\mathcal{W})$ and
$y_1^{\mathcal{W}_i},\ldots,y_L^{\mathcal{W}_i}\sim
p(y|\mathbf{x},\mathcal{D},\mathcal{W}=\mathcal{W}_i)$ for $i=1,\ldots,M$.

\subsection{Toy Problems}\
We now will illustrate the active learning procedure described in the previous section on two toy examples. In each problem we will first train a BNN with 2 layers and 20 units in each layer on the available data. Afterwards, we approximate the information-theoretic measures as outlined in the previous section

\begin{figure*}[t]
\centering
\subfloat[][]{\includegraphics[scale=0.21]{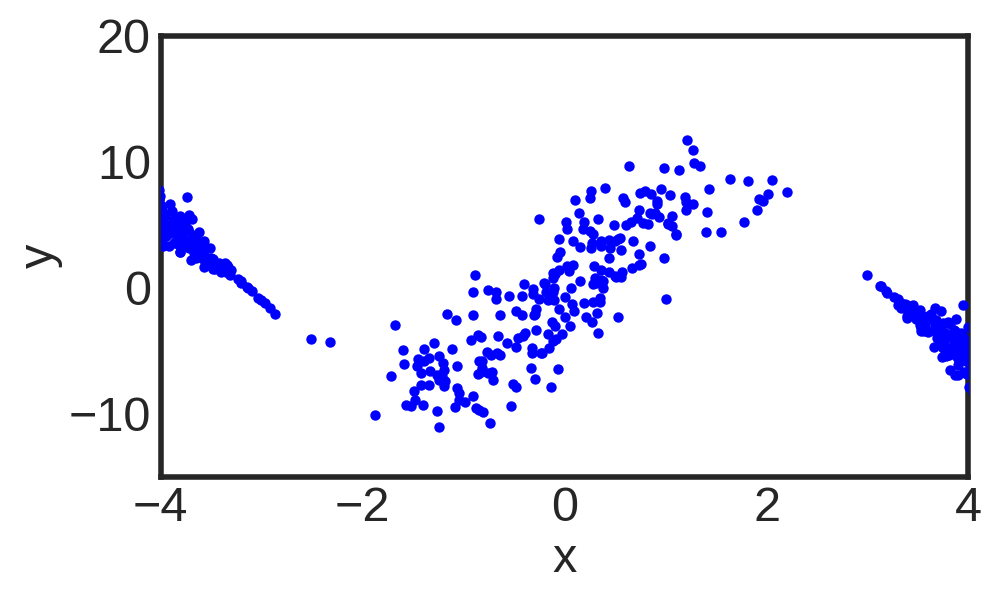}\label{fig:rd}}
\subfloat[][]{\includegraphics[scale=0.21]{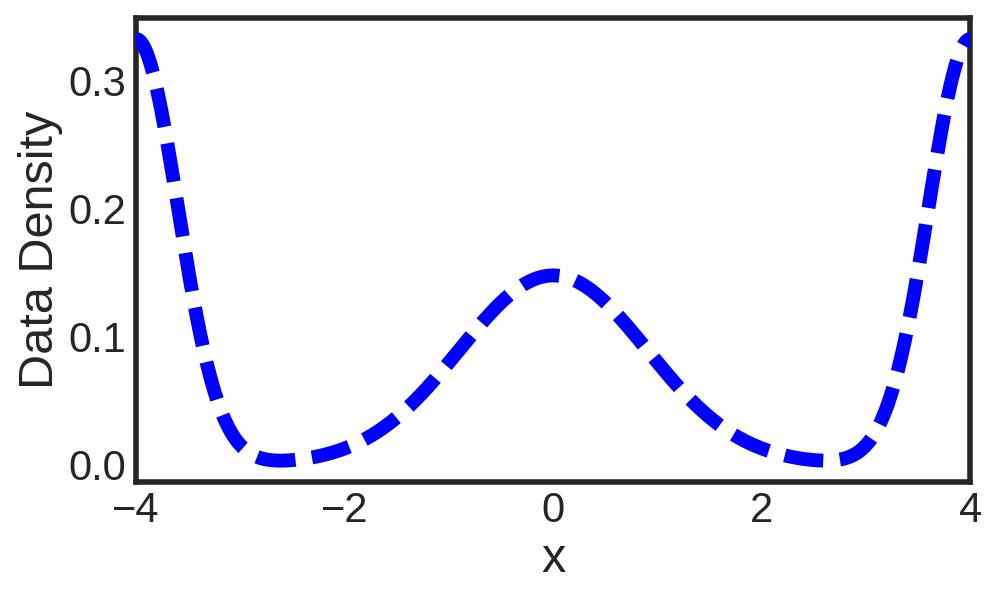}\label{fig:dd}}
\subfloat[][]{\includegraphics[scale=0.21]{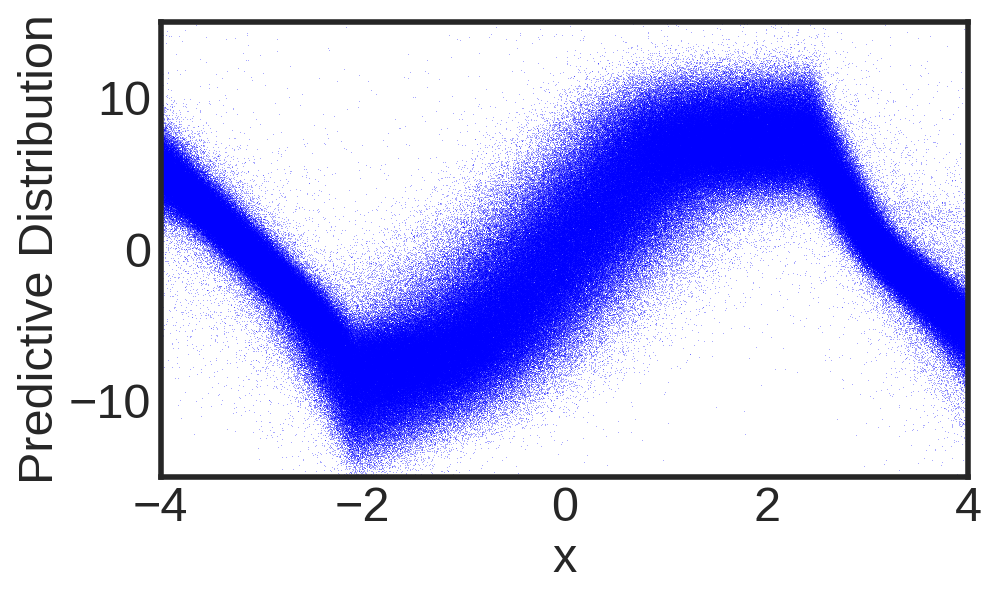}\label{fig:pd}} \\
\vspace{-0.25cm}
\subfloat[][]{\includegraphics[scale=0.21]{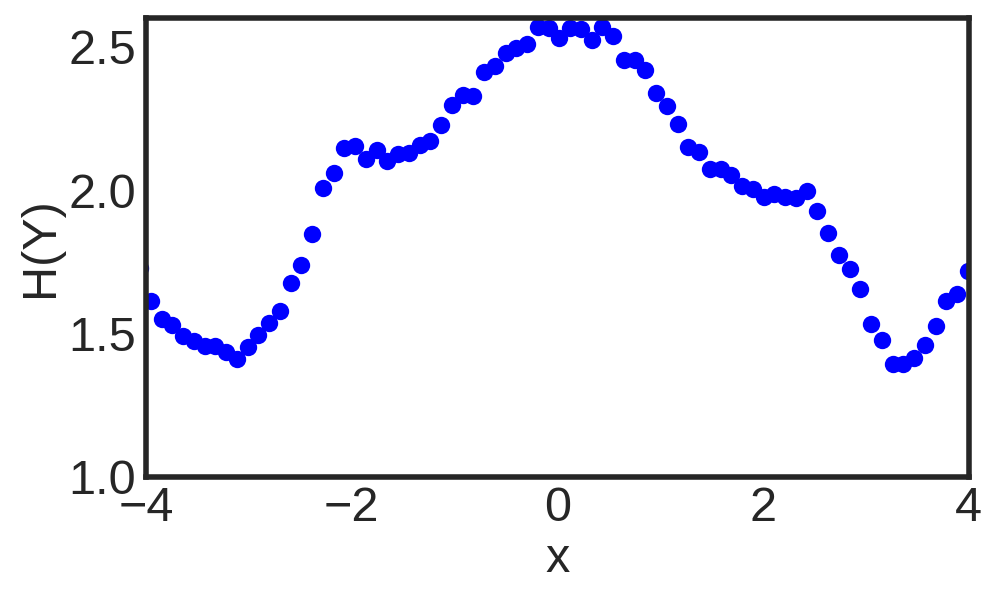}\label{fig:h}}
\subfloat[][]{\includegraphics[scale=0.21]{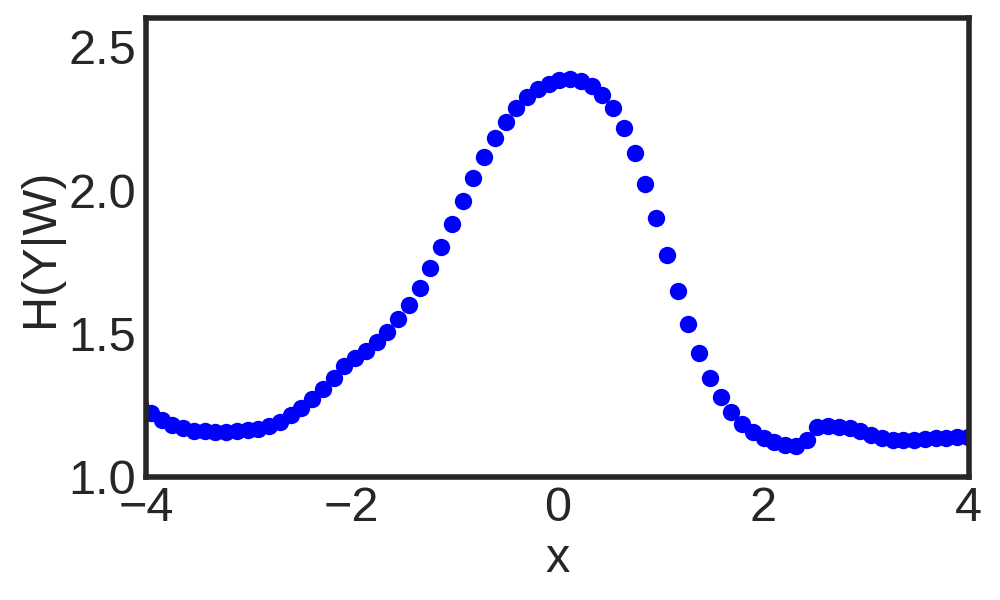}\label{fig:hgw}}
\subfloat[][]{\includegraphics[scale=0.21]{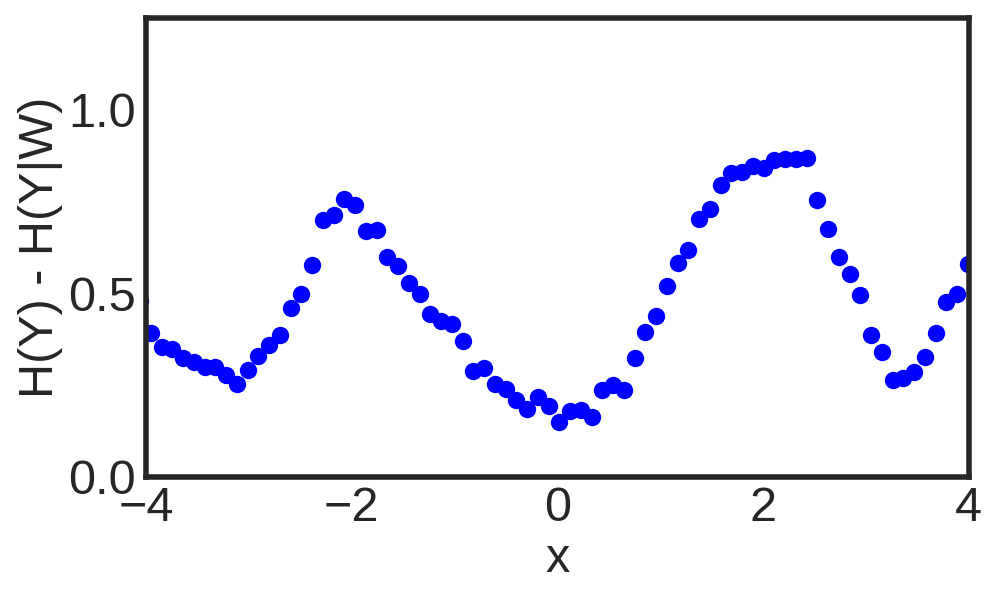}\label{fig:ob}}
\caption{Active learning example using heteroskedastic data. \protect\subref{fig:rd}: Raw data.
  \protect\subref{fig:dd}: Density of $x$ in raw data. \protect\subref{fig:pd}: Predicitive distribution: $p(y|x)$
  of BNN.  \protect\subref{fig:h}: Entropy estimate $\text{H}(y|x)$ of predictive distribution for each $x$. \protect\subref{fig:hgw}: Conditional Entropy estimate $\mathbf{E}_\mathcal{W} \text{H}(y|x,\mathcal{W})$ of predictive distribution for each $x$. \protect\subref{fig:ob}: Estimate of reduction
  in entropy for each $x$.}
  \label{toy_ep}
  \end{figure*}

We first consider a toy problem given by a regression task with heteroskedastic noise. For this, we define 
the stochastic function
$y= 7 \sin (x) + 3|\cos (x / 2)| \epsilon$ with $\epsilon \sim \mathcal{N}(0,1)$. The data availability is
limited to specific regions of $x$. In particular, we sample 750 values of $x$
from a mixture of three Gaussians with mean parameters 
$\{\mu_1= -4,\mu_2= 0,\mu_3= 4\}$, variance
parameters $\{\sigma_1=\frac{2}{5},\sigma_2=0.9,\sigma_3=\frac{2}{5}\}$ and with each Gaussian component
having weight equal to $1/3$ in the mixture.
Figure \ref{fig:rd} shows the raw data. We have
lots of points at both borders of the $x$ axis and in the center, but little data available in
between. 

\begin{figure*}[t]
\centering
\subfloat[][]{\includegraphics[scale=0.21]{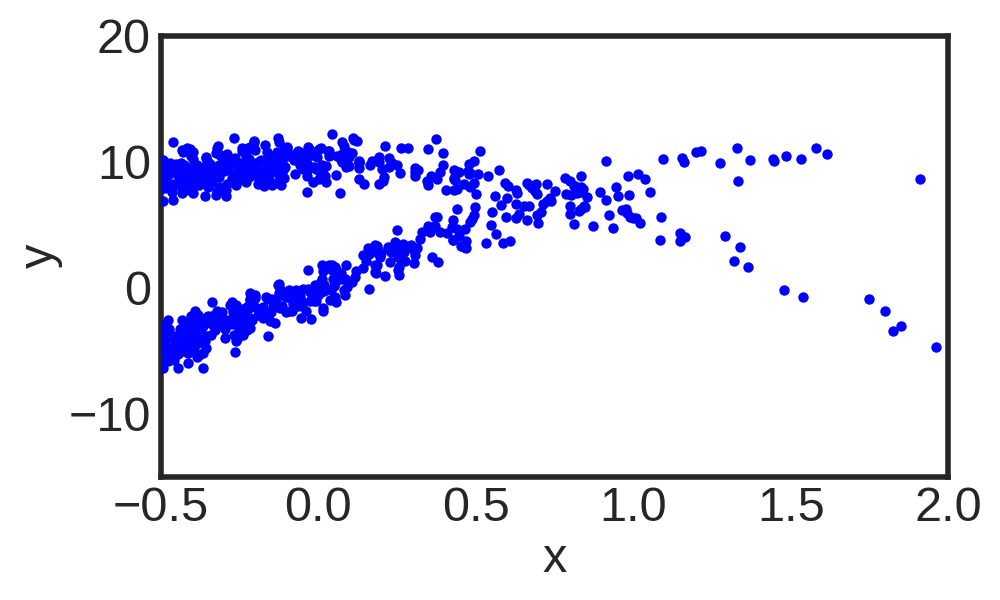}\label{fig:brd}}
\subfloat[][]{\includegraphics[scale=0.21]{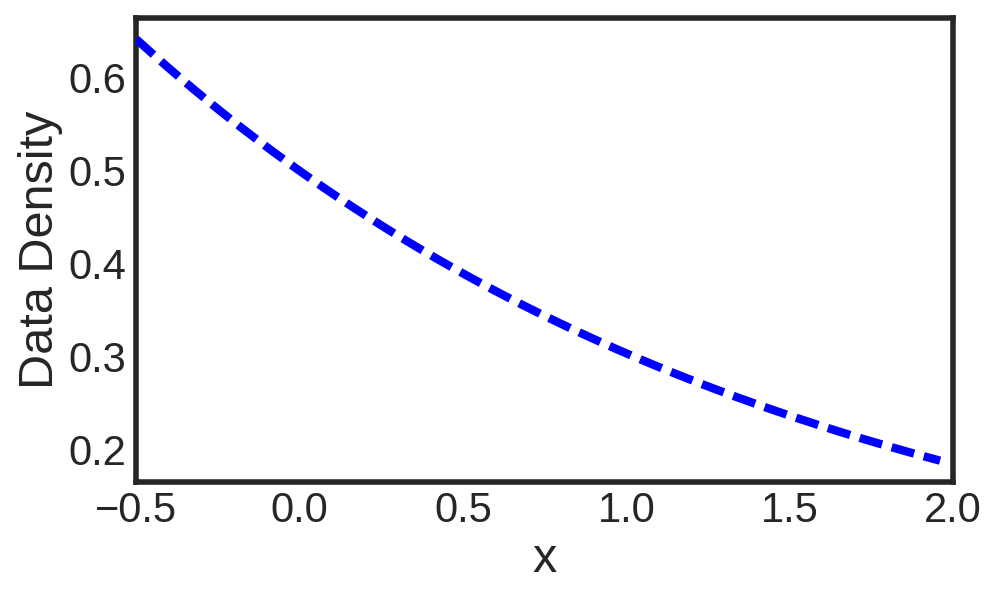}\label{fig:bdd}}
\subfloat[][]{\includegraphics[scale=0.21]{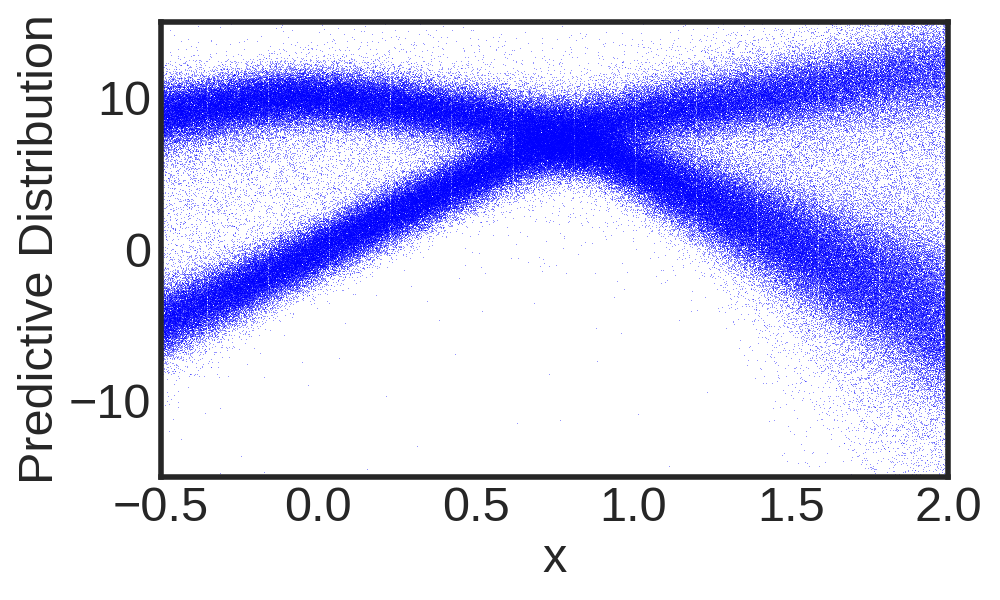}\label{fig:bpd}} \\
\vspace{-0.25cm}
\subfloat[][]{\includegraphics[scale=0.21]{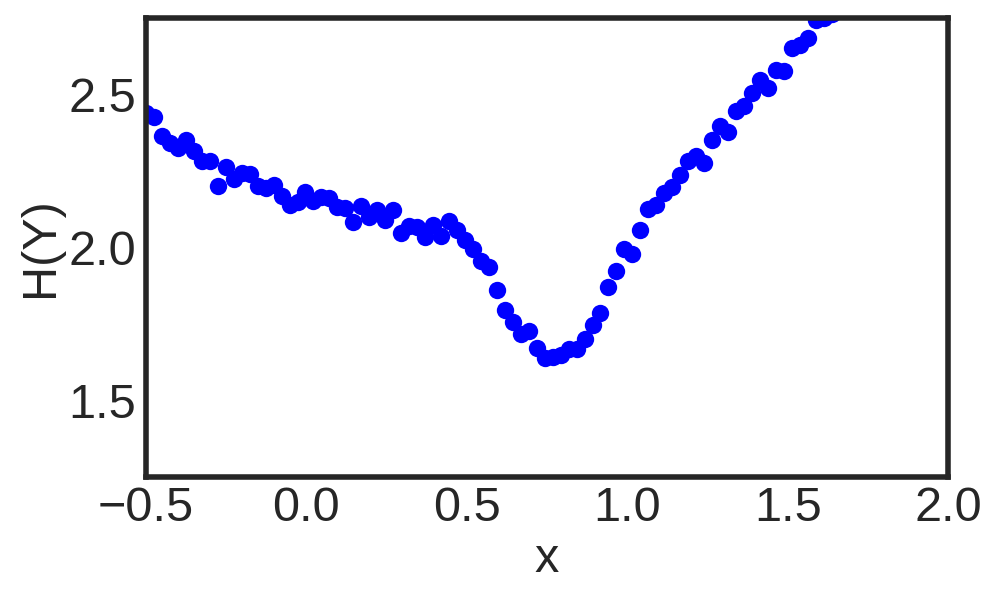}\label{fig:bh}}
\subfloat[][]{\includegraphics[scale=0.21]{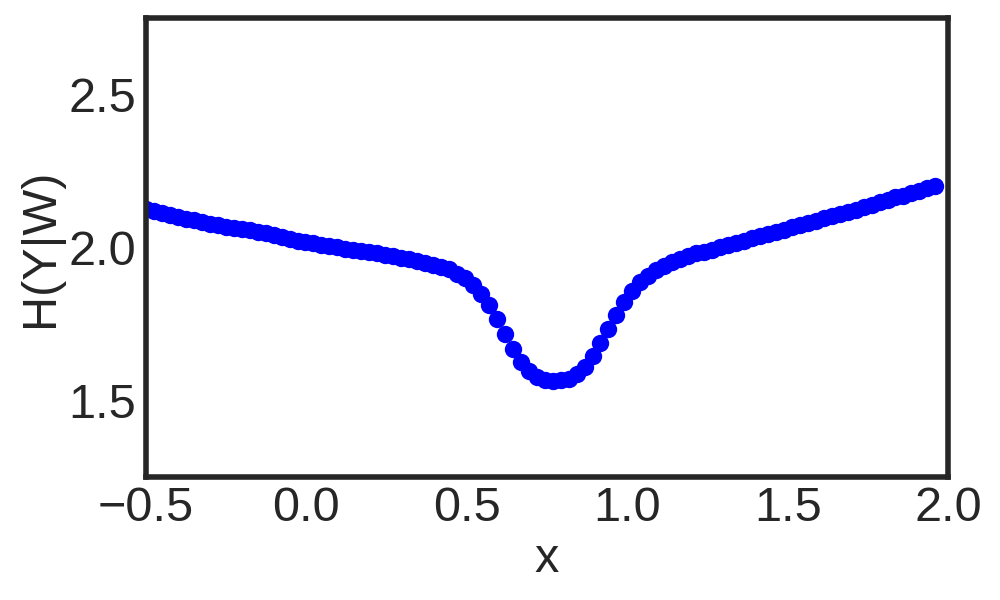}\label{fig:bhgw}}
\subfloat[][]{\includegraphics[scale=0.21]{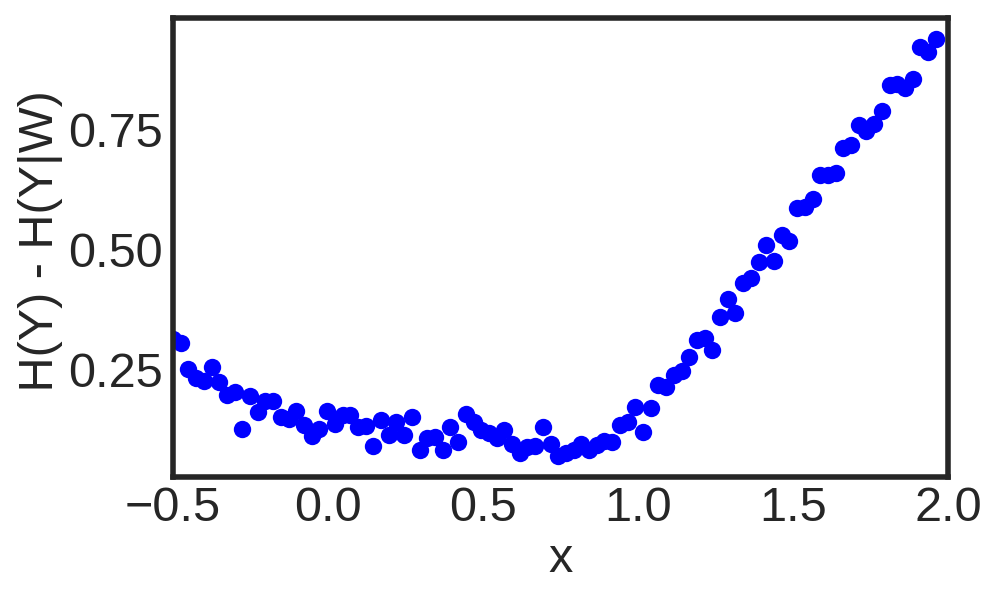}\label{fig:bob}}
\caption{Active learning example using bimodal  data. \protect\subref{fig:rd}: Raw data.
  \protect\subref{fig:dd}: Density of $x$ in raw data. \protect\subref{fig:pd}: Predicitive distribution: $p(y|x)$
  of BNN.  \protect\subref{fig:h}: Entropy estimate $\text{H}(y|x)$ of predictive distribution for each $x$. \protect\subref{fig:hgw}: Conditional Entropy estimate $\mathbf{E}_\mathcal{W} \text{H}(y|x,\mathcal{W})$ of predictive distribution for each $x$. \protect\subref{fig:ob}: Estimate of reduction
  in entropy for each $x$.}
  \label{toy_bep}
  \end{figure*}

Figure \ref{toy_ep} visualizes the respective
quantities. We see that the BNN with latent variables does an accurate
decomposition of its predictive uncertainty between epistemic uncertainty and
aleatoric uncertainty: the reduction in entropy approximation, as shown in
Figure \ref{fig:ob}, seems to be inversely proportional to the density used to
sample the data (shown in Figure \ref{fig:dd}). This makes sense, since in this
toy problem the most informative data points are expected to be located in regions
where data is scarce. Note that, in more complicated settings, the most
informative data points may not satisfy this property.

Next we consider a toy problem given by a regression task with bimodal data.
We define $x\in[-0.5, 2]$ and $y=10\sin (x)+\epsilon$
with probability $0.5$ and $y=10\cos (x)+\epsilon$, otherwise, where $\epsilon
\sim \mathcal{N}(0,1)$ and $\epsilon$ is independent of $x$. The data availability is
not uniform in  $x$. In particular we sample 750 values of  $x$ from an exponential distribution with $\lambda=0.5$

Figure \ref{toy_bep}  visualizes the respective
quantities. The predictive distribution shown in Figure \ref{fig:bpd} suggests that the BNN has learned the bimodal structure in the data. The predictive distribution appears to get increasingly 'washed out' as we increase $x$. This increase in entropy as a function of $x$ is shown  in Figure \ref{fig:bh}.  The conditional entropy $\text{H}(y|\mathcal{W})$ of the predictive distribution shown in Figure \ref{fig:bhgw} appears to be symmetric around $x=0.75$. 
This suggest that the BNN has correctly learned to separate the  aleatoric component from the full uncertainty for the problem at hand: the ground truth function is symmetric around $x=0.75$ at which point it changes from a bimodal to a unimodal stochastic function. Figure \ref{fig:bob} shows the estimate of reduction in entropy for each $x$. Here we can observe two effects: First, as expected, the expected entropy reduction will increase with higher $x$. Second, we see a slight
decrease from $x=-0.5$ to $x=0.75$.  We believe the reason for this is twofold: because the data is limited to $[-0.5,2]$ we expect a higher level of uncertainty in the vicinity of both borders. Furthermore we expect that learning a bimodal function requires more data to reach the same level of confidence than a unimodal function. 

\section{Risk-Sensitive Reinforcement Learning}

In the previoius section we studied  how BNNs with latent variables can be used
for active learning of stochastic functions. The resulting algorithm is
based on a decomposition of predictive uncertainty into its aleatoric and
epistemic components. In this section we build up on this result to derive a
new risk-sensitive objective in model-based RL with the aim to minimize the
effect of model bias. Our new risk criteron enforces that the learned
policies, when evaluated on the ground truth system, are safe in the sense that
on average they do not deviate significantly from the performance prediced by
the model at training time.

Similar to \cite{depeweg2016learning}, we consider the domain of batch
reinforcement learning. In this setting we are given a batch of state transitions
$\mathcal{D}=\{(\mathbf{s}_t, \mathbf{a}_t,\mathbf{s}_{t+1})\}$ formed by
triples containing the current state $\mathbf{s}_t$, the action applied
$\mathbf{a}_t$ and the next state $\mathbf{s}_{t+1}$. For example, $\mathcal{D}$ may be
formed by measurements taken from an already-running system. In addition to
$\mathcal{D}$, we are also given a cost function $c$. The goal is to obtain
from $\mathcal{D}$ a policy in parametric form that minimizes $c$ on average under the
system dynamics. 

The aforementioned problem can be solved using model-based policy search
methods.  These methods include two key parts \citep{deisenroth2013survey}. The
first part consists in learning a dynamics model from $\mathcal{D}$.  We assume
that the true dynamical system can be expressed by an unknown neural network
with stochastic inputs: \begin{equation}\label{eq:transitions} \mathbf{s}_t =
f_\text{true}(\mathbf{s}_{t-1},\mathbf{a}_{t-1},z_t;\mathcal{W}_\text{true})\,,\quad
z_t \sim \mathcal{N}(0,\gamma)\,, \end{equation} where
$\mathcal{W}_\text{true}$ denotes the synaptic weights of the network and
$\mathbf{s}_{t-1}$, $\mathbf{a}_{t-1}$ and $z_t$ are the inputs to the network.
In the second part of our model-based policy search algorithm, we optimize a
parametric policy given by a deterministic neural network with synaptic weights
$\mathcal{W}_{\pi}$.  
This parametric policy computes the action $\mathbf{a}_t$ as a function of $\mathbf{s}_t$, that is,
 $\mathbf{a}_t = \pi(\mathbf{s}_t;\mathcal{W}_\pi)$.
We optimize $\mathcal{W}_{\pi}$ to  minimize the expected
cost $C = \sum_{t=1}^T c_t$ over a finite horizon $T$ with respect to our
belief $q(\mathcal{W})$, where $c_t = c(\mathbf{s}_t)$. This expected cost is obtained by averaging over
multiple virtual roll-outs. For each roll-out we choose $\mathbf{s}_0$
randomly from the states in $\mathcal{D}$, sample $\mathcal{W}_i\sim q$
and then simulate state trajectories using the model
$\mathbf{s}_{t+1}=f(\mathbf{s}_t,\mathbf{a}_t,z_t;\mathcal{W}_i)+\bm
\epsilon_{t+1}$ with policy $\mathbf{a}_t = \pi(\mathbf{s}_t;\mathcal{W}_\pi)$,
input noise $z_t \sim \mathcal{N}(0,\gamma)$ and additive noise $\bm
\epsilon_{t+1} \sim \mathcal{N}(\bm 0, \bm \Sigma)$. This procedure allows us
to obtain estimates of the policy's expected cost for any particular cost
function. If model, policy and cost function are differentiable, we are then
able to tune $\mathcal{W}_{\pi}$ by stochastic gradient descent over the
roll-out average.

Given the cost function $c$, the objective to be optimized by the policy search algorithm is
\begin{align}
\textstyle J(\mathcal{W}_{\pi})  = \textstyle \mathbf{E}_{q(\mathcal{W})}\left[C\right]= \textstyle \mathbf{E}_{q(\mathcal{W})} \left[\sum_{t=1}^{T}c_t\right]\,.\label{eq:exact_cost}
\end{align}
In practice, $\mathbf{s}_0$ is sampled uniformly at random from the training data $\mathcal{D}$.
Standard approaches in risk-sensitive RL
\citep{garcia2015comprehensive,maddison2017particle,mihatsch2002risk} use
the standard deviation of the cost $C$ as risk measure. High risk is
associated to high variability in the cost $C$. To penalize risk, the new objective to be optimized is given by
\begin{align}
\textstyle J(\mathcal{W}_{\pi})  = \textstyle \mathbf{E} \left[C\right] +  
\beta \mathbf{\sigma} (C)\,,  \label{mf-risk}
\end{align}
where
$\mathbf{\sigma} (C)$ denotes the standard deviation of the cost 
and the free parameter $\beta$ determines the amount of risk-avoidance ($\beta \ge 0$) or risk-seeking
behavior ($\beta < 0$). In this standard setting, the variability of the cost
$\mathbf{\sigma} (C)$ originates from two different sources. First, from the existing uncertainty
over the model parameters and secondly, from the intrinsic stochasticity of the dynamics.


One of the main dangers of model-based RL is model bias: the discrepancy of
policy behavior under a) the assumed model and b) the ground truth MDP. While
we cannot avoid the existence of such discrepancy when data is limited, we wish
to guide the policy search towards policies that stay in state spaces where the
risk for model-bias is low. For this, we can define the model bias $b$ as follows:
\begin{equation}
b(\mathcal{W}_{\pi}) = \sum_{t=1}^T  \left|\mathbf{E}_\text{true} [c_t] - \mathbf{E}_{q(\mathcal{W})} [c_t] \right|\,,\label{eq:model_bias}
\end{equation}
where $\mathbf{E}_\text{true} [c_t]$ is the expected cost obtained at time $t$
when starting at the initial state $\mathbf{s}_0$ and the ground truth dynamics are evolved according to policy
$\pi(\mathbf{s}_t ; \mathcal{W}_{\pi})$. Note that we focus on having similar
expectations of the individual state costs $c_t$ instead of having similar
expectations of the final episode cost $C$. The former is a more strict
criterion since it may occur that model and ground truth diverge,
but both give roughly the same cost $C$ on average.

As indicated in (\ref{eq:transitions}), we assume that the true dynamics are
determined by a neural network with latent variables and weights given
by $\mathcal{W}_\text{true}$. By using the approximate posterior
$q(\mathcal{W})$, and assuming that $\mathcal{W}_\text{true}\sim
q(\mathcal{W})$, we can obtain an upper bound on the expected model bias
as follows:

\vspace{-0.4cm}
\resizebox{\linewidth}{!}{
  \begin{minipage}{\linewidth}
\small
\begin{align}
\mathbf{E}_{q(\mathcal{W})}&[b(\mathcal{W}_{\pi}] =  \mathbf{E}_{\mathcal{W}_\text{true} \sim q(\mathcal{W})}   
\sum_{t=1}^T  \left|\mathbf{E}[c_t|\mathcal{W}_\text{true}] - \mathbf{E}_{q(\mathcal{W})} [c_t] \right| \nonumber \\ 
 &=   \sum_{t=1}^T  \mathbf{E}_{\mathcal{W}_\text{true} \sim q(\mathcal{W})}  \sqrt{(\mathbf{E}[c_t|\mathcal{W}_\text{true}] - 
\mathbf{E}_{q(\mathcal{W})} [c_t])^2} \nonumber \\ 
 &\le     \sum_{t=1}^T   \sqrt{ \mathbf{E}_{\mathcal{W}_\text{true} \sim q(\mathcal{W})} (\mathbf{E}[c_t|\mathcal{W}_\text{true}] - 
\mathbf{E}_{q(\mathcal{W})} [c_t])^2} \nonumber\\
 &= \sum_{t=1}^T   \sqrt{ \sigma_{\mathcal{W}_\text{true} \sim q(\mathcal{W})}^2(\mathbf{E}[c_t|\mathcal{W}])} \nonumber \\
 &=    \sum_{t=1}^T   \sigma_{\mathcal{W}_\text{true} \sim q(\mathcal{W})}(\mathbf{E}[c_t|\mathcal{W}_\text{true}])\,. \label{eq:risk_rl}
\end{align} 
\end{minipage}}

\vspace{0.1cm}
We note that $\mathbf{E}[c_t|\mathcal{W}]$ is the expected reward of a policy $\mathcal{W}_{\pi}$ under the dynamics
given by $\mathcal{W}$. The expectation integrates out the influence of the
latent variables $z_1,\ldots,z_t$ and the output noise $\epsilon_1,\ldots,\epsilon_t$. The last equation in (\ref{eq:risk_rl}) can thereby be
interpreted as the variability of the reward, that originates from our
uncertainty over the dynamics given by distribution $q(\mathcal{W})$. 

In Section \ref{sec:al} we showed how (\ref{eq:al-objective}) encodes
decomposition of the entropy of the predictive distribution 
into its aleatoric and epistemic components. The resulting
decomposition naturally arises from an information-theoretic approach for active learning.
We can express $\sigma^2_{\mathcal{W}_\text{true} \sim q(\mathcal{W})}(\mathbf{E}[c_t|\mathcal{W}_\text{true}])$
in a similar way using the law of total variance: 

\vspace{-0.5cm}
{\small
\begin{align}
\sigma^2_{\mathcal{W}_\text{true} \sim q(\mathcal{W})}(\mathbf{E}[c_t|\mathcal{W}_\text{true}]) =
\sigma^2(c_t) - \mathbf{E}_{\mathcal{W} \sim q(\mathcal{W})}[\sigma^2(c_t|\mathcal{W}])]\,.\nonumber
\end{align}
}We extend the policy search objective of (\ref{eq:exact_cost}) with a risk
component given by an approximation to the model bias.  Similar to
\citet{depeweg2016learning}, we derive a Monte Carlo approximation that enables
optimization by gradient descent. For this, we perform $M \times N$
roll-outs by first sampling $\mathcal{W}\sim q(\mathcal{W})$ a total of $M$ times and
then, for each of these samples of $\mathcal{W}$, performing $N$ roll-outs in
which $\mathcal{W}$ is fixed and we only sample the latent variables and the
additive Gaussian noise. In particular,

\vspace{-0.2cm}
\resizebox{\linewidth}{!}{
  \begin{minipage}{\linewidth}
\footnotesize
\begin{align}
 &J(\mathcal{W}_{\pi})  = \sum_{t=1}^T \left\{ \mathbf{E}_{q(\mathcal{W})}\left[c_t\right] +  \beta 
\sigma_{\mathcal{W}_\text{true} \sim q(\mathcal{W})}(\mathbf{E}[c_t|\mathcal{W}_\text{true}])
\right\} 
\approx \nonumber\\
 &  \sum_{t=1}^{T} \left\{ \frac{1}{MN } \left[ \sum_{m=1}^M \sum_{n=1}^N  c_{m,n}(t) \right] +
\beta \hat{\sigma}_{M}\left( \frac{1}{N}\sum_{n=1}^{N} c_{m,n}(t)\right) \right\}\,,\label{eq:risk_rl1} 
\end{align}
\vspace{-0.05cm}
\end{minipage}}
where $c_{m,n}(t) = c(\mathbf{s}_{t}^{\mathcal{W}^{m},\{z_{1}^{m,n},\ldots,z_{t}^{m,n}\}, \{\bm{\epsilon}_{1}^{m,n},\ldots,\bm{\epsilon}_{t}^{m,n}\},\mathcal{W}_\pi})$ is the cost that is obtained at time $t$ in a roll-out
generated by using a policy with parameters $\mathcal{W}_\pi$, a transition function parameterized by
$\mathcal{W}^m$ and latent variable values $z_1^{m,n},\ldots,z_{t}^{m,n}$, with additive noise values
$\bm{\epsilon}_{1}^{m,n},\ldots,\bm{\epsilon}_{t}^{m,n}$. $\hat{\sigma}_{M}$ is an empirical estimate of the standard deviation calculated
over $M$ draws of $\mathcal{W}$.

The free parameter $\beta$ determines the importance of the risk criterion. As
described above, the proposed approximation generates $M \times N$ roll-out
trajectories for each starting state $\mathbf{s}_0$. For this, we sample
$\mathcal{W}^m \sim q(\mathcal{W})$ for $m=1,\ldots,M$ and for each $m$ we then
do $N$ roll-outs with different draws of the latent variables $z_t^{m,n}$ and
the additive Gaussian noise $\epsilon_t^{m,n}$. We average across the $M\times
N$ roll-outs to estimate $\mathbf{E}_{\mathcal{W} \sim
q(\mathcal{W})}[c_t]$. Similarly, for each $m$, we average across the
corresponding $N$ roll-outs to estimate 
$\mathbf{E}[c_t|\mathcal{W}^m]$. Finally, we compute the empirical standard
deviation of the resulting estimates to approximate
$\sigma_{\mathcal{W}_\text{true} \sim q(\mathcal{W})}(\mathbf{E}[c_t|\mathcal{W}_\text{true}])$.



\begin{figure*}[th!]
\centering
\subfloat[][]{\includegraphics[scale=0.375]{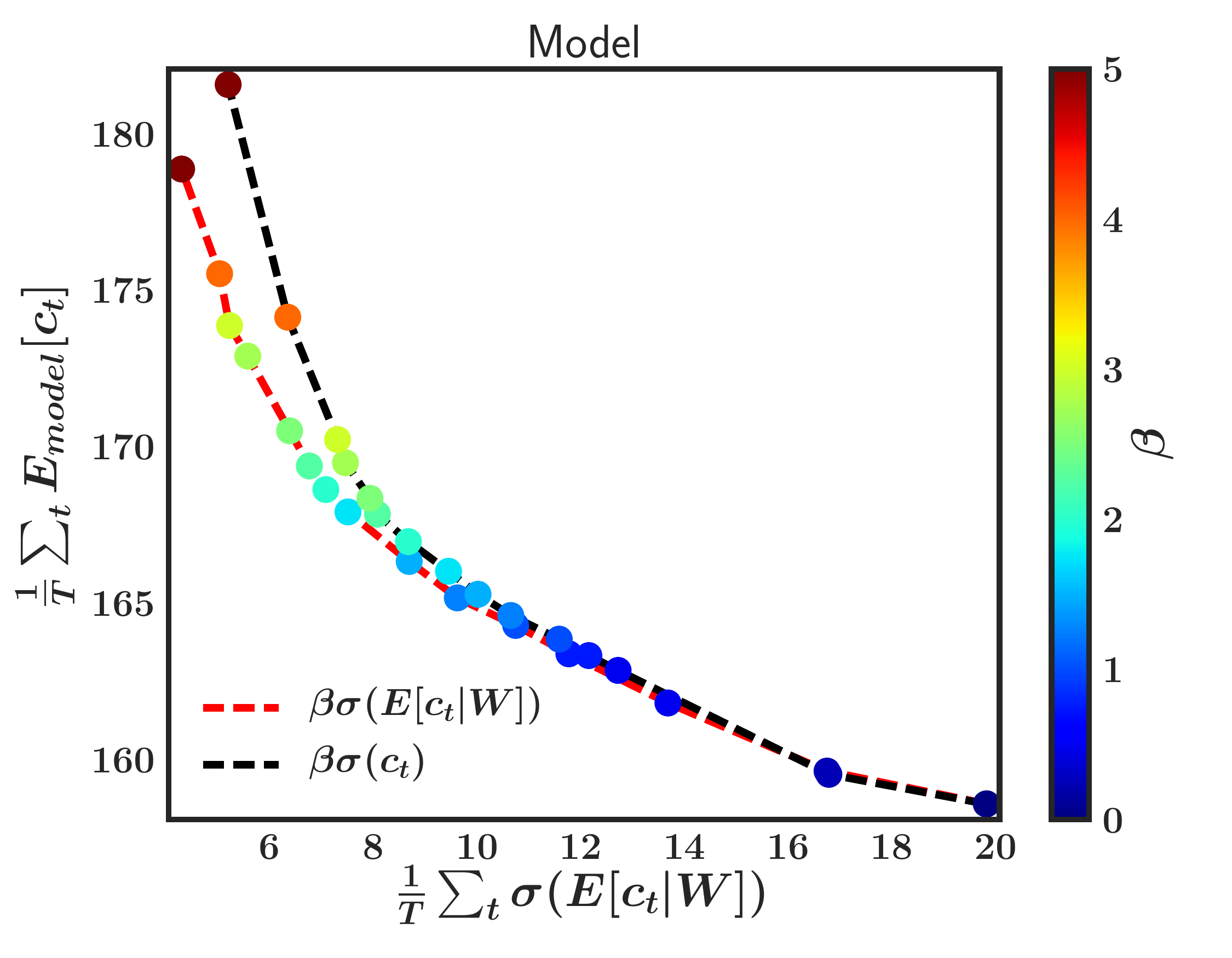}\label{fig:ids_train}}
\subfloat[][]{\includegraphics[scale=0.375]{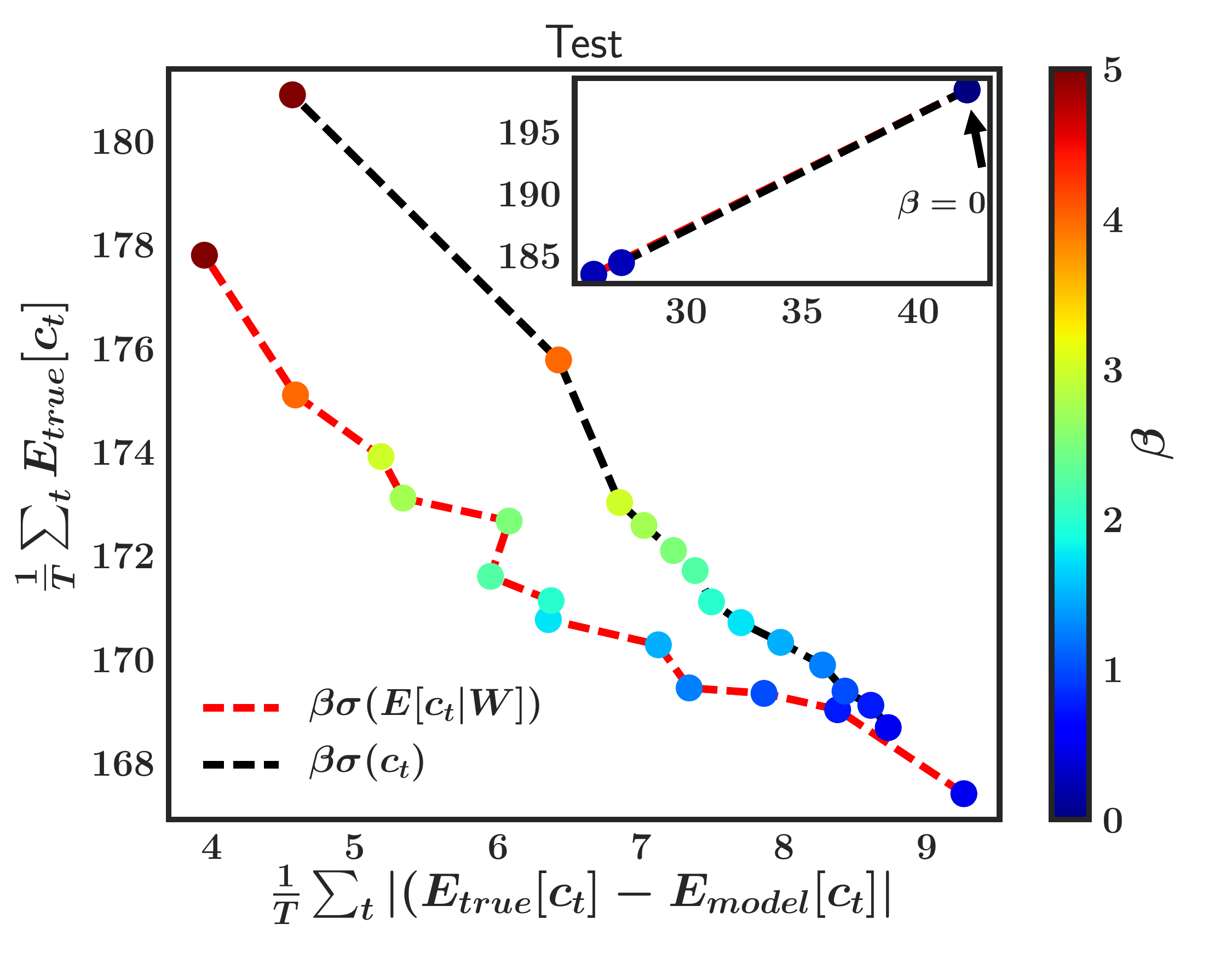}\label{fig:ids_test}}

\caption{Results on Industrial Benchmark. Performances 
of policies trained using equation (\protect\ref{eq:risk_rl1}) (red curve) and baseline that
minimizes (\ref{mf-risk}) (black curve) for different values of $\beta$. Figure
\protect\subref{fig:ids_train} shows results under the model and Figure
\protect\subref{fig:ids_train} shows results under the ground truth.
}
  \label{ids_result}
  \end{figure*}
  
\subsection{Application: Industrial Benchmark}

We show now the effectiveness of the proposed method on a stochastic dynamical system. For
this we use the industrial benchmark, a high-dimensional stochastic model
inspired by properties of real industrial systems. A detailed description and
example experiments can be found in
\cite{hein2016introduction,depeweg2016learning}, with python source code
available\footnote{\url{https://github.com/siemens/industrialbenchmark}}\footnote{\url{https://github.com/siemens/policy_search_bb-alpha}}.

In our experiments, we first define a behavior policy
that is used to collect data by interacting with the system. This policy is
used to perform three roll-outs of length $1000$ for each setpoint value in
$\{0,10,20,\ldots,100\}$. The setpoint is a hyper-parameter of the industrial
benchmark that indicates the complexity of its dynamics. The setpoint is
included in the state vector $\mathbf{s}_t$ as a non-controllable variable which
is constant throughout the roll-outs. Policies in the industrial benchmark specify
changes $\Delta_v$, $\Delta_g$ and $\Delta_s$ in three steering variables $v$ (velocity), $g$ (gain)
and $s$ (shift) as a function of $\mathbf{s}_t$. In the behavior
policy these changes are stochastic and sampled according to
\begin{align}
\Delta_v & \sim  \left\{
\begin{array}{@{}ll@{}}
    \mathcal{N}(0.5,\frac{1}{\sqrt{3}})\,, & \text{if}\,\, v(t) < 40 \\
    \mathcal{N}(-0.5,\frac{1}{\sqrt{3}})\,, & \text{if}\,\, v(t) > 60 \\
    \mathcal{U}(-1,1)\,, & \text{otherwise}
  \end{array}
  \right.\\
\Delta_g & \sim  \left\{
\begin{array}{@{}ll@{}}
    \mathcal{N}(0.5,\frac{1}{\sqrt{3}})\,, & \text{if}\,\, g(t) < 40 \\
    \mathcal{N}(-0.5,\frac{1}{\sqrt{3}})\,, & \text{if}\,\, g(t) > 60 \\
    \mathcal{U}(-1,1)\,, & \text{otherwise}
  \end{array}
  \right.\\
\Delta_s & \sim  \mathcal{U}(-1,1) \,.
\end{align}
The velocity $v(t)$ and gain $g(t)$ can take values in $[0,100]$. Therefore,
the data collection policy will try to keep these values only in the medium
range given by the interval $[40,60]$. Because of this, large parts of the
state space will be unobserved. After collecting the data, the 
$30,000$ state transitions are used to train a BNN with latent variables
with the same hyperparameters as in \cite{depeweg2016learning}.

After this, we train different policies using the Monte Carlo approximation
described in equation (\ref{eq:risk_rl1}). We consider different choices of
$\beta\in [0,5]$ and use a horizon of $T=100$ steps, with $M=50$ and $N=25$ and
a minibatch size of $1$. 

Performance is measured using two different objectives. The first one is the
expected cost obtained under the ground truth dynamics of the system, that is
$\sum_{t=1}^T \mathbf{E}_\text{true} [c_t]$. The second objective is the model
bias as defined in equation (\ref{eq:model_bias}).  
We compare with two baselines. The first one ignores any risk and, therefore,
is obtained by just optimizing equation (\ref{eq:exact_cost}). The second
baseline uses the standard deviation $\sigma(c_t)$ as risk criterion and, therefore,
is similar to equation (\ref{mf-risk}), which is the
standard approach in risk-sensitive RL. 

In Figure \ref{ids_result} we show the results obtained by our method and
by the second baseline when performance is evaluated under the model (Figure
\ref{fig:ids_train}) or under the ground truth (Figure \ref{fig:ids_test}).
Each plot shows empirical estimates of the model bias vs. the expected
cost, for various choices of $\beta$. We also highlight the result obtained
with $\beta=0$, the first baseline.

Our novel approach for risk-sensitive reinforcement learning produces policies
that attain at test time better trade-offs between expected cost and model
bias. As $\beta$ increases, the policies gradually put more emphasis on the
expected model bias. This leads to higher costs but lower discrepancy between
model and real-world performance.

\section{Conclusion}

We have studied a decomposition of predictive uncertainty into its epistemic and aleatoric
components when working with Bayesian neural networks with
latent variables. This decomposition naturally arises in an
information-theoretic active learning setting. The decomposition also inspired
us to derive a novel risk objective for safe reinforcement learning that
minimizes the effect of model bias in stochastic dynamical systems.

\bibliography{references}
\bibliographystyle{icml2017}
\end{document}